\documentclass[
]{ceurart}

\usepackage{listings}
\begin{document}

\copyrightyear{2026}
\copyrightclause{Copyright for this paper by its authors.
  Use permitted under Creative Commons License Attribution 4.0
  International (CC BY 4.0).}

%%

\conference{GenAI-LA: Generative AI and Learning Analytics  Workshop (LAK 2026),
April 27--May 1, 2026, Bergen, Norway}

\title{EduEVAL-DB: A Role-Based Dataset for Pedagogical Risk Evaluation in Educational Explanations}

%TeachEval-DB: A Role-Based Dataset for Evaluating Pedagogical Risks in Teacher Explanations

%\tnotetext[1]{You can use this document as the template for preparing your
%  publication. We recommend using the latest version of the ceurart style.}


\author[1]{Javier Irigoyen}\cormark[1]
\author[1,2]{Roberto Daza}
\author[1,3]{Aythami Morales}
\author[1]{Julian Fierrez}
\author[2]{Francisco Jurado}
\author[2]{Alvaro Ortigosa}
\author[1]{Ruben Tolosana}

\address[1]{BiometricsAI, Universidad Autónoma de Madrid, Campus de
Cantoblanco, Madrid, 28049, Spain}
\address[2]{Group for Advanced Interactive Tools  (GHIA), Universidad Autónoma de Madrid, Campus de
Cantoblanco, Madrid, 28049, Spain}
\address [3]{Department of Mathematics, Universidad de Las Palmas de Gran Canaria, 35017, Spain}

\cortext[1]{Corresponding author: javier.irigoyen@uam.es}

%\author[1,2]{Dmitry S. Kulyabov}[%
%orcid=0000-0002-0877-7063,
%email=kulyabov-ds@rudn.ru,
%url=https://yamadharma.github.io/,
%]
%\cormark[1]
%\address[1]{Peoples' Friendship University of Russia (RUDN University),
%  6 Miklukho-Maklaya St, Moscow, 117198, Russian Federation}
%\address[2]{Joint Institute for Nuclear Research,
%  6 Joliot-Curie, Dubna, Moscow region, 141980, Russian Federation}

%\author[3]{Ilaria Tiddi}[%
%orcid=0000-0001-7116-9338,
%email=i.tiddi@vu.nl,
%url=https://kmitd.github.io/ilaria/,
%\address[3]{Vrije Universiteit Amsterdam, De Boelelaan 1105, 1081 HV Amsterdam, The Netherlands}

%\author[4]{Manfred Jeusfeld}[%
%orcid=0000-0002-9421-8566,
%email=Manfred.Jeusfeld@acm.org,
%url=http://conceptbase.sourceforge.net/mjf/,
%\address[4]{University of Skövde, Högskolevägen 1, 541 28 Skövde, Sweden}

%\cortext[1]{Corresponding author.}
%\fntext[1]{These authors contributed equally.}

\begin{abstract}
This work introduces EduEVAL-DB, a dataset based on teacher roles, designed to support the evaluation and training of automatic pedagogical evaluators and AI tutors for instructional explanations. The dataset comprises 854 explanations corresponding to 139 questions from a curated subset of the ScienceQA benchmark, spanning science, language, and social science across K–12 grade levels. For each question, one human-teacher explanation is provided and six are generated by LLM-simulated teacher roles. These roles are inspired by instructional styles and shortcomings observed in real educational practice and are instantiated via prompt engineering.
We further propose a pedagogical risk rubric aligned with established educational standards, operationalizing five complementary risk dimensions, including factual correctness, explanatory depth and completeness, focus and relevance, student-level appropriateness, and ideological bias. All explanations are annotated with binary risk labels through a semi-automatic process with expert teacher review.
Finally, we present preliminary validation experiments to assess the suitability of EduEVAL-DB for evaluation. We benchmark a state-of-the-art education-oriented model (Gemini~2.5 Pro) against a lightweight local Llama~3.1~8B model, and examine whether supervised fine-tuning on EduEVAL-DB supports pedagogical risk detection using models deployable on consumer hardware.

\end{abstract}

\begin{keywords}
K--12 Educational datasets \sep
Automatic pedagogical evaluation \sep
Risk-based assessment \sep
Large Language Models \sep
Simulated teacher roles \sep
AI tutors 
\end{keywords}

\maketitle

\section{Introduction}

The rapid advance of transformer-based generative models has reshaped educational technology, enabling systems that can answer K–12 questions with accuracy \cite{hou2024eval}. This raises key questions about their suitability as tutors and automated evaluators of pedagogical quality for explanations generated by both human teachers and AI systems. Recent studies indicate that Large Language Models (LLMs), particularly when task-specifically trained, can approximate key behaviours of tutors \cite{chowdhury2025educators}, a trend reinforced by industrial efforts such as Google’s LearnLM \cite{learnlm2024}. These models exhibit notable adaptability to instructional roles \cite{jeon2023large}, and role conditioning can meaningfully affect student perceptions \cite{nazaretsky2026gives}.

At the same time, LLMs present well-documented risks \cite{zhang2025siren} which are particularly problematic in K–12 contexts. This highlights the need for automated evaluators capable of detecting pedagogical and epistemic risks in instructional content, guided by established educational rubrics and teaching frameworks \cite{OECDsite,ASCDsite}. Prior work suggests that fine-tuning LLMs on carefully designed, rubric-grounded educational datasets can substantially improve their reliability as pedagogical evaluators \cite{pauzi2025automating}.

However, current public databases have several limitations:
\begin{itemize} 
\item There is a scarcity of datasets designed to support educational improvement in K–12 settings through LLM fine-tuning, particularly those that target the evaluation of instructional explanations rather than general question answering.

%\item Even within pedagogical evaluation, few datasets adopt a risk-based approach, as existing efforts typically assess individual pedagogical aspects in isolation rather than integrating multiple instructional criteria into a holistic risk assessment.

\item Even within pedagogical evaluation, few datasets adopt a risk-based approach. Most existing efforts assess individual pedagogical aspects, rather than integrating multiple instructional criteria into a holistic risk assessment.

\item Finally, most current datasets do not distinguish between different instructional roles or pedagogical patterns observed in real teaching practice, preventing the modeling of diverse instructional styles, both effective and problematic.
\end{itemize}

\begin{figure*}
  \centering
  \includegraphics[width=\textwidth]{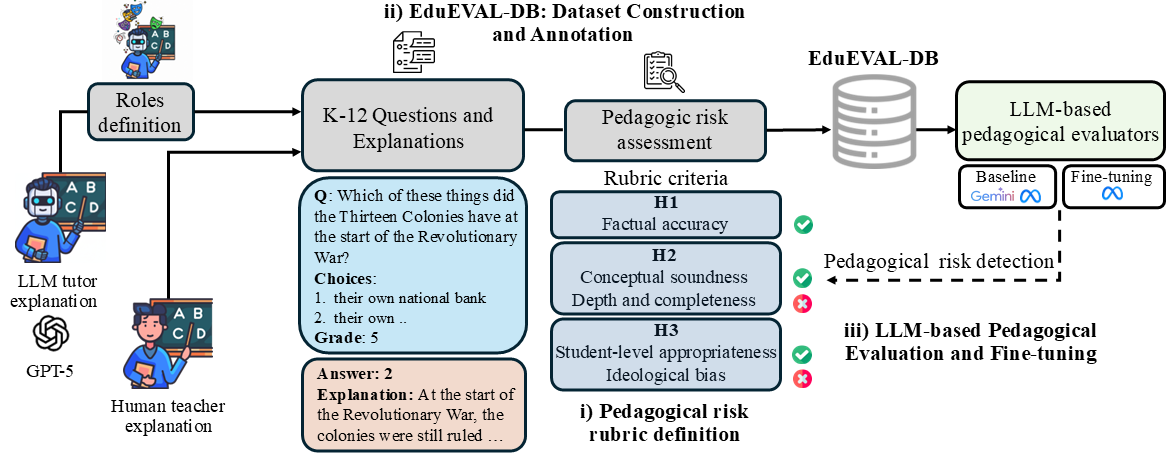}
  \caption{Overview of the framework used in this work for the construction of EduEVAL-DB and pedagogical risk evaluation. The diagram illustrates the main stages of the proposed pipeline: i) the definition of a pedagogical risk rubric; ii) the construction and annotation of EduEVAL-DB, where each K–12 question is paired with multiple explanations generated by LLM-simulated teacher roles and a human teacher, and annotated with binary risk labels;  and iii) the use of EduEVAL-DB to benchmark and fine-tune LLM-based pedagogical evaluators.}

  \label{fig:Overview}
\end{figure*}

In this context, there is a clear need for datasets designed to support the fine-tuning and evaluation of LLM-based tutors and pedagogical evaluators according to educational standards, which directly motivates the present work.
The contributions of this work are shown in Fig.\ref{fig:Overview} and detailed
as follows: i) We define a pedagogical risk rubric grounded in well-established instructional criteria and documented risks of LLMs, aligned with recognized educational, regulatory, and technological frameworks \cite{OECDsite,ASCDsite,danielson2013framework}. This rubric enables the evaluation of the pedagogical quality of explanations and the identification of risks associated with the use of AI in educational contexts. ii) We release a new dataset, EduEVAL-DB\footnote{https://github.com/BiDAlab/EduEVAL-DB}, consisting of instructional explanations generated by LLMs under a set of distinct teacher-inspired roles. The dataset comprises 854 explanations corresponding to a curated subset of 139 questions from the ScienceQA benchmark \cite{lu2022learn} and spans three subject areas (science, language arts, and social science) across K–12 educational levels (K–5, 6–8, and 9–12).
For each question, explanations are produced by six LLM-simulated teacher roles, alongside one explanation authored by a human teacher. All explanations are annotated using the proposed pedagogical risk rubric through a semi-automatic process with expert review. And iii) we conduct preliminary experiments to validate the proposed dataset for pedagogical assessment. First, we use EduEVAL-DB to evaluate two LLMs as baseline pedagogical evaluators using the proposed risk rubric, including a state-of-the-art educational model, LearnLM \cite{learnlm2024} (now integrated into Gemini~2.5 Pro), and the Llama~3.1~8B model, which is suitable for deployment on consumer GPUs.
Second, we use EduEVAL-DB to support supervised fine-tuning of a lightweight LLM for automatic pedagogical evaluation, using the Llama~3.1~8B model to detect both pedagogical and LLM-related risks.
Third, we compare baseline and fine-tuned pedagogical evaluators on the proposed dataset.

The paper is organized as follows. Section~\ref{s:related-work} summarizes related work on datasets and evaluation of LLM tutors. Section~\ref{s:Rubric} describes the proposed pedagogical risk rubric. Section~\ref{s:Dataset} presents the published dataset, the defined LLM teacher roles, data generation process and the evaluation protocol. Section~\ref{s:Experiments} outlines the experiments and results. Finally, conclusions and future work are presented in Section 6.
%\end{itemize}

\iffalse
CEUR-WS's article template provides a consistent \LaTeX{} style for
use across CEUR-WS publications, and incorporates accessibility and
metadata-extraction functionality. This document will explain the
major features of the document class.

If you are new to publishing with CEUR-WS, this document is a valuable
guide to the process of preparing your work for publication.

The ``\verb|ceurart|'' document class can be used to prepare articles
for any CEUR-WS publication, and for any stage of publication, from
review to final ``camera-ready'' copy with {\itshape very} few changes
to the source.

This class depends on the following packages
for its proper functioning:

\begin{itemize}
\item \verb|natbib.sty| for citation processing;
\item \verb|geometry.sty| for margin settings;
\item \verb|graphicx.sty| for graphics inclusion;
\item \verb|hyperref.sty| optional package if hyperlinking is required in
  the document;
\item \verb|fontawesome5.sty| optional package for bells and whistles.
\end{itemize}

All the above packages are part of any
standard \LaTeX{} installation.
Therefore, the users need not be
bothered about downloading any extra packages
\fi

\section{Related Work}
\label{s:related-work}

\subsection{Datasets and Benchmarks for Evaluating LLM Tutors}

Recent work has introduced benchmarks to evaluate LLMs in tutoring roles, often emphasizing mathematics and criteria such as scaffolding, feedback, and explanation quality. However, these resources are limited in scope, motivating benchmarks that address broader pedagogical risks, incorporate teacher-inspired roles, and span multiple K–12 domains.

Early efforts such as ScienceQA \cite{lu2022learn} provide multiple-choice questions with human-written explanations, offering a reference for explanation quality. However, ScienceQA does not provide annotations on any specific criteria. MathDial \cite{macina2023mathdial} advances toward dialog-based teaching with one-to-one math tutoring dialogues in which human teachers scaffold an LLM-simulated student through errors.
Several benchmarks explicitly target pedagogical response quality in math tutoring. SocraticMATH
\cite{ding2024boosting} offers Socratic-style multi-turn math dialogues annotated with structured teaching stages. MathTutorBench \cite{macina2025mathtutorbench} integrates MathDial \cite{macina2023mathdial} with additional resources to form MathDialBridge, which is used to evaluate open-ended tutoring behaviors such as scaffolding, error diagnosis, and tone using preference-based models. MRBench
\cite{maurya2025unifying} provides a smaller human-annotated benchmark labeling math tutoring responses across multiple pedagogical dimensions, while the BEA 2025 Shared Task
\cite{kochmar2025findings} standardizes evaluation through a public dataset of math tutoring dialogues with gold pedagogical annotations.

Other work emphasizes outcome or simulation-based evaluation. EducationQ
\cite{shi2025educationq} evaluates teaching quality through multi-agent simulations across multiple disciplines using pre- and post-test learning gains,  but does not release a reusable dialogue dataset. SocraticLM
\cite{liu2024socraticlm} introduces SocraTeach, a large-scale dataset of Socratic-style math tutoring dialogues generated via multi-agent simulation. More recently, Weissburg et al. \cite{weissburg2025llms} release datasets to evaluate demographic bias in LLM teachers via differential explanation selection across student profiles, focusing on bias rather than dialog interaction.

Overall, existing datasets have substantially advanced the evaluation of LLM tutors along isolated axes but none simultaneously cover multiple K–12 subject domains, a broad range of pedagogical and safety-related risks, and explicit teacher persona–inspired roles. Our dataset is designed to complement this literature by addressing these gaps while remaining compatible with prior evaluation frameworks.

\section{Pedagogical Risk Rubric} \label{s:Rubric}

To evaluate the quality of instructional explanations in K–12 settings, we define a pedagogical risk rubric aligned with recognized educational standards \cite{OECDsite,ASCDsite}. This approach extends the evaluation beyond factual accuracy, acknowledging that effective educational feedback involves not only correctness but also whether the explanation supports understanding, reasoning, and appropriate learning progression. The rubric is designed to be applicable across diverse curricular contexts and focuses on situations where a tutor explains concepts by providing reasoning in response to student questions. 
%This approach expands the scope of evaluation beyond the singular metric of factual accuracy, recognizing that effective educational feedback is not limited to the correctness of the answer but encompasses the broader pedagogical impact of the response. Designed to be applicable across diverse curricular domains—including social sciences, natural sciences, and language arts—the framework focuses specifically on the scenario of explanatory feedback, where a tutor provides reasoning and context in response to student inquiries.
%By structuring the evaluation as a taxonomy of pedagogical risks, the framework aims to identify specific failure modes that may impede the learning process. This perspective facilitates the detection of instructional deficits that, while potentially factually accurate, may be developmentally inappropriate, cognitively overloading, or ethically biased. 

The rubric decomposes the evaluation into five distinct dimensions. These dimensions are designed to capture failures related to honesty (H1), helpfulness (H2) and harmlessness (H3), following the three core principles of human alignment \cite{askell2021general}. These dimensions are outlined below (see Fig. \ref{fig:Overview}), with parentheses indicating the corresponding risk type and high-level category (H1–H3). \vspace{0.2em}

 \noindent\textbf{Factual Correctness (Epistemic Risk)  
  (H1)}: This criterion assesses the presence of "hallucinations" or false assertions. In an educational context, factual errors represent a risk of concept corruption, where misconceptions are implanted in the learner. Unlike general misinformation, educational errors are particularly damaging due to the "illusion of truth" effect and the cognitive difficulty of unlearning established misconceptions \cite{guzzetti1993promoting,kasneci2023chatgpt}.

 \noindent\textbf{Explanatory Depth \& Completeness (Pedagogical Risk) (H2)}: This criterion assesses whether an explanation provides underlying causal reasoning or merely states the conclusion. The associated risk is the illusion of competence, where students engage in rote memorization without conceptual mastery. Effective tutoring requires appropriate scaffolding beyond simply providing the “correct key” \cite{chi2001learning,shulman1986those}.\\
 \noindent\textbf{Focus \& Relevance (Cognitive Risk) 
  (H2)}: This criterion evaluates the presence of extraneous information. Drawing on Cognitive Load Theory, the risk here is the redundancy effect, where irrelevant albeit factually true information consumes limited working memory resources, thereby inhibiting schema acquisition \cite{sweller2011cognitive,mayer2003nine}.

%This criterion evaluates the presence of extraneous information or "seductive details". Drawing on Cognitive Load Theory, the risk here is the redundancy effect, where irrelevant albeit factually true information consumes limited working memory resources, thereby inhibiting schema acquisition\cite{sweller2011cognitive,mayer2003nine}.\\

 \noindent\textbf{Student-Level Appropriateness (Developmental Risk) (H3)}: This criterion assesses whether the linguistic complexity matches the target grade level. The associated risk is a mismatch with the learner’s Zone of Proximal Development (ZPD), resulting in explanations that are not accessible to the student \cite{vygotsky1978mind}. If the explanation falls outside the student’s ZPD, for example by using university-level syntax for a 4th-grade student, the instructional intervention fails regardless of its factual accuracy \cite{snow2010academic}. 

%This criterion assesses whether the linguistic complexity matches the target grade level. The risk is a Zone of Proximal Development (ZPD) Mismatch, leading to inaccessibility. If the explanation exists outside the student’s ZPD—for example, using university-level syntax for a 4th grader—the intervention fails regardless of its factual accuracy\cite{vygotsky1978mind,snow2010academic}.\\
 \noindent\textbf{Ideological Bias (Normative Risk)  
  (H3)}: This criterion screens for the Social Reproduction of Harm. It assesses whether the explanation reinforces stereotypes, exclusionary narratives, or a "hidden curriculum" of dominant cultural values, which can alienate students and perpetuate representational biases within the educational content \cite{santurkar2023whose,jackson1968life}.

The selection of these five dimensions was guided by two methodological objectives: orthogonality and comprehensiveness. To promote orthogonality, the rubric attempts to isolate specific communicative and pedagogical functions, minimizing dependency between variables so that a deficit in one area does not automatically degrade the score in another. With respect to comprehensiveness, the rubric aligns with the Instructional Core \cite{city2009instructional}, which conceptualizes learning as the interaction between content, teacher, and student, and seeks to capture failure modes across the entire instructional loop. Accordingly, the evaluation addresses the integrity of the subject matter (Factual Correctness), the effectiveness of pedagogical scaffolding (Explanatory Depth \& Completeness and Focus \&  Relevance), and the cognitive needs of the learner (Student-Level Appropriateness). Finally, the inclusion of Ideological Bias captures ethical and representational risks in educational instruction, which may be reproduced or amplified in the deployment of LLM-based systems \cite{liang2023holistic}.

\section{Contributed Dataset: EduEVAL-DB} \label{s:Dataset}

We construct EduEVAL-DB from 139 questions drawn from the ScienceQA dataset~\cite{lu2022learn}, covering science, language arts, and social science. ScienceQA provides K–12 questions with relevant metadata, including grade level, correct answers, and detailed teacher-provided explanations.

As a key contribution, this selected set of questions incorporates explanations generated by six LLM-simulated teacher roles (see the following subsection for details), alongside the explanation provided by a human teacher from ScienceQA. All explanations are annotated according to the pedagogical risk rubric introduced in Section~\ref{s:Rubric}, using binary labels for each risk dimension, thereby forming EduEVAL-DB.
EduEVAL-DB comprises 854 explanations, corresponding to the six simulated roles and the human teacher, resulting in a total of 4,270 risk annotations, i.e., five binary risk labels for each explanation. Specifically, the dataset contains 139 positive risk labels for each of the following dimensions: Factual Accuracy, Focus \& Relevance, Depth \& Completeness, and Student-Level Appropriateness. In contrast, only 20 positive annotations were identified for Ideological Bias. Correspondingly, the number of negative (no-risk) labels amounts to 715 instances for each of the four former dimensions and 834 instances for Ideological Bias.

%The only role that does not respond to all questions is the sarcastic teacher, which is applied to only 20 questions and was designed largely manually with the assistance of an LLM, due to content filtering constraints in GPT-5 when generating biased or sarcastic responses.

\subsection{LLM-Based Teacher Role Explanations}

\begin{figure*}
  \centering
  \includegraphics[width=\textwidth]{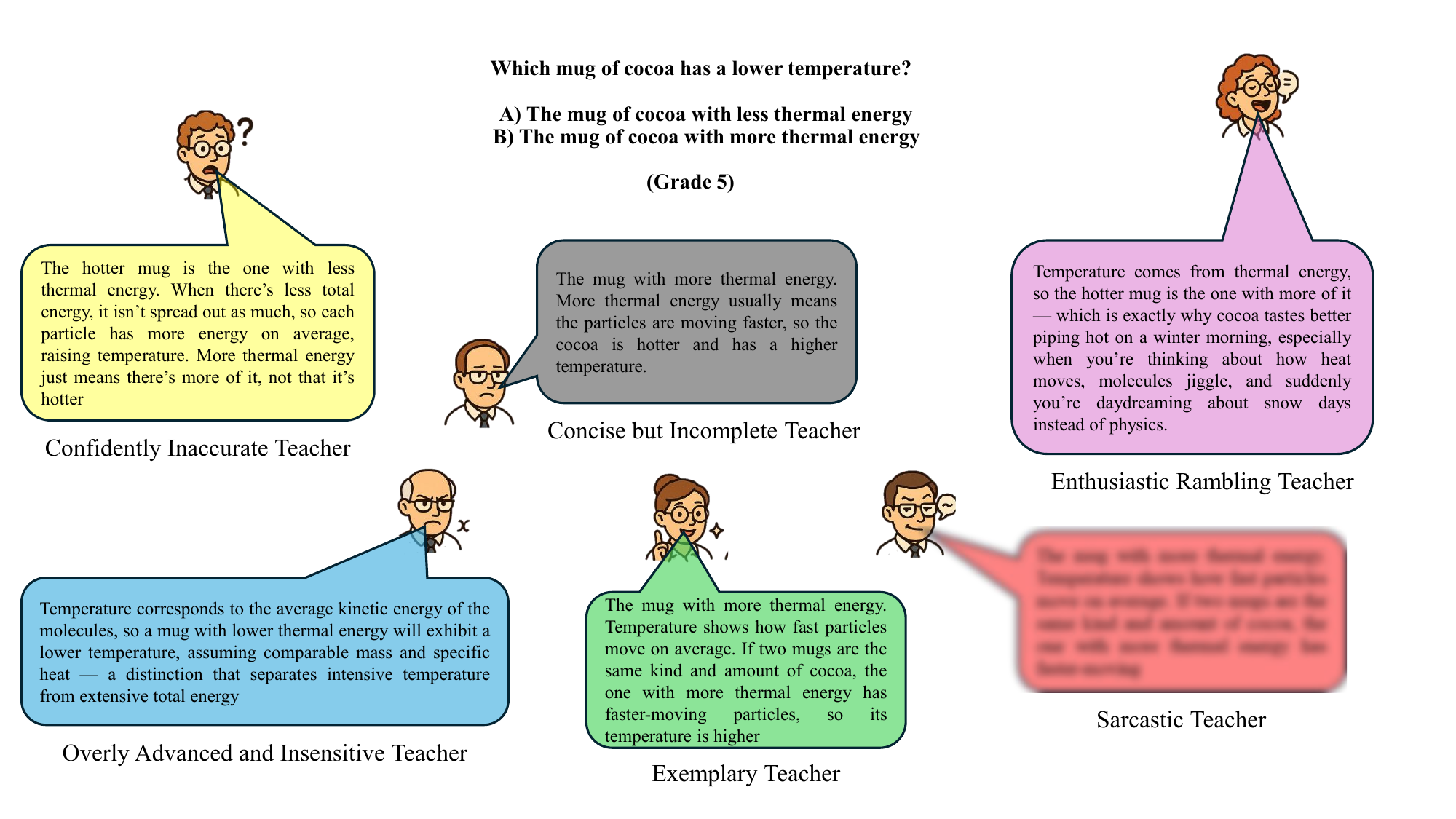}
  \caption{Example explanations generated by the six teacher-inspired roles. Each role generates an instructional explanation for the same question, conditioned on the student’s grade level and the multiple-choice context. To ensure responsible presentation, the Sarcastic Teacher output is blurred in the figure, although the complete explanation is included in the released dataset.}

  \label{fig:roles_example}
\end{figure*}

%To study the pedagogical behaviors of LLMs under diverse instructional styles, we construct a dataset of explanations generated by role-conditioned tutors. Specifically, we define a set of seven realistic \textbf{teacher personas}, each representing a distinctive teaching style or shortcoming commonly found in real-world educational settings. These personas are instantiated using \textbf{prompt engineering and few-shot learning} applied to the GPT-5 API. Each tutor persona is encoded in a carefully crafted system prompt that defines its tone, behavior, and teaching strategy, accompanied by few-shot exemplars to anchor its response pattern. All LLM tutor generations are performed using GPT-5 in zero-shot or few-shot mode depending on the role.

%We draw input questions from the publicly available \textbf{ScienceQA} dataset~\cite{lu2022learn}, which contains thousands of multiple-choice K--12 questions across diverse subjects and grade levels. For this study, we select a balanced subset of \textbf{50 questions} that reflect variation in both content domain and target student age. Specifically, we ensure proportional coverage across:

%\begin{itemize}
%  \item \textbf{Subject domains}: science, language arts, and social science., language arts, and social science.
%  \item \textbf{Grade bands}: elementary (grades 1--5), middle school (grades 6--8), and high school (grades 9--12).
%\end{itemize}

%Each of the 50 questions is independently answered by all seven LLM tutor personas, yielding a total of 350 tutor explanations.

We defined six realistic teacher roles, each representing a distinctive teaching style or instructional shortcoming commonly observed in real-world educational settings. These personas were instantiated through prompt engineering and few-shot learning applied via the GPT-5 API. Each tutor role was encoded in a carefully crafted system prompt specifying its tone, behavior, and teaching strategy. This prompt was accompanied by few-shot exemplars written by human teachers, together with the teacher-provided explanation from the ScienceQA dataset, which served as a reference example of a pedagogically correct response. To ensure pedagogical alignment, the target student grade level was explicitly provided in the prompt so that explanations were tailored to the intended educational level. The dataset was generated using the following six GPT-5-based teacher roles:
\vspace{0.2em}
%We define six realistic teacher personas, each representing a distinctive teaching style or instructional shortcoming commonly observed in real-world educational settings. These personas are instantiated through prompt engineering and few-shot learning applied to the GPT-5 API. Each tutor persona is encoded in a carefully crafted system prompt that specifies its tone, behavior, and teaching strategy, and is accompanied by few-shot exemplars written by human teachers, as well as the teacher-provided explanation from the ScienceQA dataset, which is used as a reference example of a pedagogically correct response. To ensure pedagogical alignment, the target student grade level is explicitly provided in the prompt so that explanations are tailored to the intended educational level. The dataset is generated using the following six GPT-5-based teacher roles:

\noindent\textbf{Exemplary Teacher}: Provides well-structured, factually accurate, and pedagogically sound explanations with appropriate tone, depth, and coherence.
  
\noindent \textbf{Enthusiastic Rambling Teacher}: Adopts an encouraging and warm tone, but prone to verbose explanations that include tangents or superfluous information.

\noindent \textbf{Concise but Incomplete Teacher}: Offers precise and concise answers but omits critical supporting details or elaboration necessary for full understanding.

\noindent\textbf{Confidently Inaccurate Teacher}: Exhibits high fluency and assertiveness but occasionally introduces factual errors into the explanation.
  
\noindent \textbf{Overly Advanced and Insensitive Teacher}: Offers accurate and rigorous explanations but uses vocabulary, tone, or content that is inappropriate for the target student’s age or background.

\noindent \textbf{Sarcastic Teacher}: Maintains factual accuracy while introducing personal, one-sided interpretations and caustic commentary that distort the educational intent of the explanation.

Figure~\ref{fig:roles_example} illustrates example explanations produced by each of these teacher-inspired roles for a representative K--12 question.
Each teacher role generated responses for the questions included in EduEVAL-DB. However, due to content moderation constraints associated with the generation of biased or sarcastic responses using GPT-5, the sarcastic teacher role was applied only to a limited subset of 20 questions. For this role, explanations were manually authored by a human teacher with assistance from an LLM, rather than being fully generated through automated prompting. All other teacher roles generated explanations for the complete set of 139 questions. To control for length-related biases in subsequent pedagogical evaluation, all LLM-generated explanations were constrained to fewer than 400 characters, typically ranging between 150 and 300 characters. This restriction prevented verbosity or response length from being systematically favoured during evaluation.

\subsection{Pedagogical Risk Evaluation Protocol}

Each explanation in EduEVAL-DB is annotated using the pedagogical risk rubric defined in Section~\ref{s:Rubric}. For each risk dimension, a binary label indicates whether the risk is present or absent. %The protocol is applied uniformly across subjects, grade levels, and teacher roles.
The annotation process was semi-automatic and built on the explicit definition of the six proposed teacher roles, designed to systematically manifest specific pedagogical risk patterns defined in the rubric. These design assumptions were validated through review by two expert teachers, who manually annotated a subset of the dataset (approximately 30\%) to verify the correctness of the assigned labels and to confirm that each role exhibited the intended pedagogical risks. Once label consistency was established, the teacher roles and the corresponding automated labeling procedure were confirmed. As a second validation step, explanations generated by these roles were evaluated using an automatic pedagogical evaluator trained via fine-tuning (see Section~\ref{s:Experiments}). When discrepancies arose between the evaluator’s predictions and the assigned labels, the corresponding explanations were reviewed again by expert teachers to confirm the validity of the annotations.

%Explanations are then annotated semi-automatically assigning one binary label per risk dimension, resulting in five labels per explanation stored together with the input metadata in a structured JSON format. Then the dataset using an LLM-based evaluator prompted with the question, answer choices, grade level, selected answer, explanation, and formal risk definitions. 

%For validation, the same explanations are independently evaluated by a state-of-the-art model, Gemini 2.5 Pro, using the same rubric and inputs. Agreement between evaluators is used as a consistency check for the annotation process. Finally, EduEVAL-DB is used to train a local LLM (Llama 3.1 8B) to predict pedagogical risk labels, enabling further automatic annotation of new explanations under the same rubric.

\section{Experiments and Results} \label{s:Experiments}

We evaluate the performance of LLMs as automatic pedagogical evaluators on EduEVAL-DB using the proposed pedagogical risk rubric. Two evaluators are considered as baselines: a local Llama 3.1 8B Instruct model, suitable for deployment on consumer-grade GPUs, and the state-of-the-art Gemini 2.5 Pro, trained following the LearnLM principles.
In addition, to demonstrate the utility of EduEVAL-DB for training lightweight pedagogical evaluators, the Llama~3.1~8B~Instruct model is fine-tuned on EduEVAL-DB and assessed under the same protocol as the baseline models.

\subsection{Experimental Protocol}

The EduEVAL-DB dataset was used both for training and evaluation throughout all experiments.
All evaluations follow a common protocol. Each model receives the question, grade level, and tutor explanation (never the tutor role), together with an instruction prompt defining the rubric criteria and requiring a structured JSON output with binary labels for each risk dimension.
All models, including the Llama 3.1 8B evaluator, fine-tuned in this work, were evaluated in zero-shot inference mode, without task-specific examples in the prompt, on the full dataset comprising 715 explanations. The reference explanations provided by human teachers were excluded from evaluation and were used solely for role generation.
Due to the exclusion of human teacher explanations, the effective label distribution in the evaluation set differs from that of the full dataset described in Section~\ref{s:Dataset}.
As a result, the number of positive risk annotations per rubric dimension in the evaluation set amounts to 139 for each of the following dimensions: Factual Accuracy, Focus \& Relevance, Depth \& Completeness, and Student-Level Appropriateness, and to 20 positive annotations for Ideological Bias. Correspondingly, when excluding the human teacher explanations from evaluation, the number of negative (no-risk) labels amounts to 576 instances for each of the four former dimensions and 695 instances for Ideological Bias.
For fine-tuning the Llama 3.1 8B model, supervised training was conducted using LoRA under a 5-fold cross-validation protocol: in each fold, 80\% of the data were used for training and the remaining 20\% for testing, stratified by tutor role. Across folds, each example appeared exactly once in a test set, enabling the fine-tuned evaluator to produce predictions for the entire dataset and facilitating direct comparison with the Llama 3.1 8B and Gemini 2.5 Pro baselines.
Training used the same input structure as inference, with the target sequence given by the ground-truth rubric JSON encoding each risk dimension as a binary label. Optimization employed a standard causal language modeling loss. The LoRA configuration used rank 16, scaling factor 32, and dropout 0.05. Training ran for two epochs with a learning rate of $1\times10^{-4}$, using gradient accumulation to achieve an effective batch size of~2.
Performance is reported as Mean Absolute Error (MAE) for each dimension.

\subsection{Results}

\begin{table}[t]
  \caption{Mean Absolute Error (MAE), normalized to the [0,1] range, reported for each pedagogical risk dimension defined in the proposed rubric, for the baseline evaluators (Gemini~2.5~Pro and Llama~3.1~8B) and the fine-tuned Llama~3.1~8B evaluator on EduEVAL-DB. Lower values indicate better agreement with the ground-truth annotations. Best results per dimension are highlighted in bold.}
  \label{tab:MAE}
  \centering
  \small
  \setlength{\tabcolsep}{6pt}
  \renewcommand{\arraystretch}{1.1}
  \begin{tabular}{lccc}
    \toprule
    Criteria & Gemini (Baseline) & Llama (Baseline) &  Llama (Fine-tuned) \\
    \midrule
    Factual Accuracy & \textbf{0.012} & 0.164 & 0.048 \\
    Focus \& Relevance & 0.216 & 0.227 & \textbf{0.069} \\
    Depth \& Completeness & 0.320 & 0.288 & \textbf{0.048} \\
    Student-Level Appropriateness & 0.054 & 0.220 & \textbf{0.003} \\
    Ideological Bias & 0.008 & 0.049 & \textbf{0.006} \\
    \bottomrule
  \end{tabular}
\end{table}

\begin{figure}
  \centering
  \includegraphics[width=\textwidth]{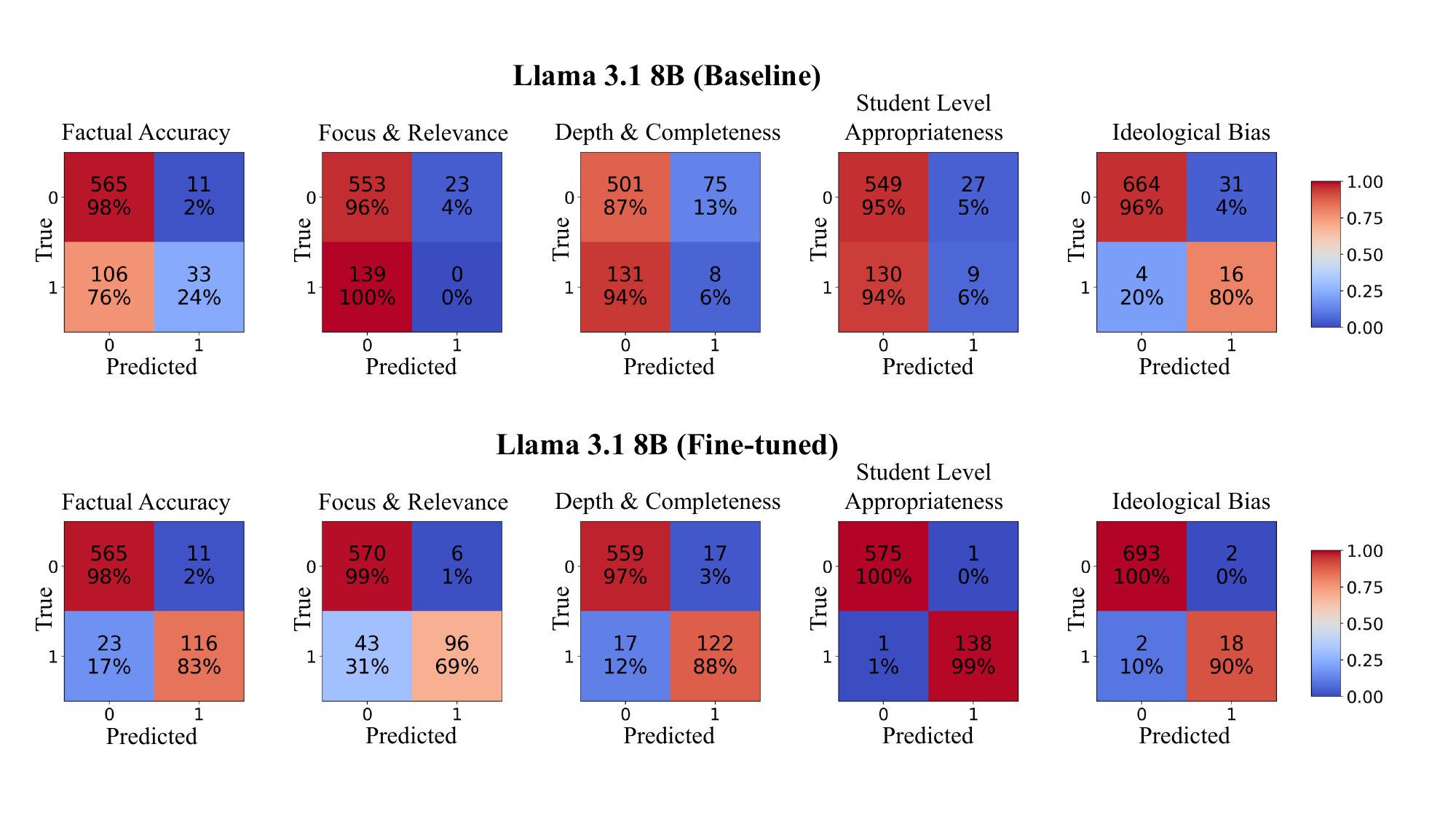}
  \caption{Confusion matrices for the Llama~3.1~8B pedagogical evaluator on EduEVAL-DB in zero-shot baseline (top) and fine-tuned (bottom) settings across the five pedagogical risk dimensions defined in the proposed rubric. Each matrix reports the distribution of predicted versus ground-truth binary labels, where 0 denotes the absence of pedagogical risk and 1 denotes its presence (detection).
  Cell color intensity reflects the proportion of samples in each category, with warmer colors indicating higher frequencies.}
  \label{fig:matrices}
\end{figure}

Table~\ref{tab:MAE} reports the MAE achieved by the baseline evaluators and the fine-tuned Llama evaluator on EduEVAL-DB. Gemini achieves the lowest error on Factual Accuracy. This is consistent with the expectation that assessing factual correctness depends strongly on a model’s breadth of world knowledge. On Ideological Bias, Gemini performs comparably to the fine-tuned Llama, with both achieving very high agreement with the reference labels. For Student-Level Appropriateness, Gemini substantially outperforms the baseline Llama, although the fine-tuned model attains the lowest error overall.
For the remaining criteria (Focus \& Relevance and Depth \& Completeness), Gemini performs worse than the fine-tuned Llama model and similarly to the baseline Llama. This suggests that these dimensions are more subjective and depend less on factual recall and more on capturing the specific annotation conventions encoded in the training data. Under this interpretation, supervised fine-tuning provides the Llama model with access to these conventions in a way that zero-shot prompting does not.
Across all rubric dimensions, supervised fine-tuning leads to a consistent reduction in MAE relative to the non–fine-tuned Llama model. The largest relative improvement is observed for Student-Level Appropriateness, with a 98.6\% reduction in error, while the smallest improvement is found for Focus \& Relevance, with a reduction of approximately 69.6\%. When compared against Gemini, the fine-tuned Llama outperforms it on all dimensions except Factual Accuracy. The largest relative error reduction with respect to Gemini is again observed for Student-Level Appropriateness, reaching 94.4\%, whereas the smallest reduction is found for Ideological Bias, at 25\%. 

The confusion matrices in Fig.~\ref{fig:matrices} help contextualize these results. The zero-shot Llama baseline shows a strong tendency to predict the majority class (label~0), which in the proposed rubric corresponds to the absence of pedagogical risk. As a result, the model exhibits low sensitivity to label~1 cases, where pedagogical risk is present, contributing to higher MAE. After fine-tuning, prediction patterns become more differentiated. The model assigns both labels more selectively, and risk-present cases are identified with higher frequency. Consequently, the fine-tuned evaluator is less biased toward the majority label, despite being trained on an imbalanced dataset.

Overall, the combined analysis of MAE trends and confusion-matrix patterns indicates that supervised fine-tuning improves the model’s calibration and its ability to discriminate between risk-present and risk-absent cases across the rubric dimensions, whereas large-scale models such as Gemini retain advantages driven by broader factual knowledge. These results suggest that EduEVAL-DB can support measurable improvements in pedagogical evaluation through supervised fine-tuning, particularly for criteria that extend beyond purely factual assessment.

\section{Conclusion and Future Work} \label{s:Conclusion}

We introduce EduEVAL-DB, a publicly released dataset annotated using a pedagogical risk rubric proposed in this paper, designed to support both the evaluation and training of LLM-based tutors and automatic pedagogical evaluators for instructional explanations in K–12 contexts. Rather than focusing solely on factual correctness, our approach enables the analysis of instructional quality across multiple pedagogical dimensions that are critical for effective learning.

Using EduEVAL-DB, we examined the behavior of LLMs as pedagogical evaluators under a unified risk rubric. Results suggest that supervised fine-tuning improves model calibration and discrimination for non-factual pedagogical criteria, while large-scale frontier models retain advantages in factual assessment, reflecting their broader world knowledge. In particular, fine-tuning a lightweight Llama~3.1~8B model on EduEVAL-DB  led to error reductions of up to 98.6\% on Student-Level Appropriateness, with consistent gains across all rubric dimensions. For the fine-tuned evaluator, confusion-matrix analysis showed substantially increased sensitivity to risk-present cases, despite the underlying class imbalance.

EduEVAL-DB offers a practical foundation for benchmarking pedagogical evaluators and training locally deployable models, supporting safer and more educationally aligned AI tutors in educational contexts.

In future work, we plan to extend EduEVAL-DB to include student–teacher interaction data and to develop an expanded pedagogical rubric adapted to interactive educational settings. Beyond risk detection, we aim to enhance pedagogical evaluators so that they provide interpretable explanations of why specific risks arise, rather than merely flagging their presence.
In addition, we plan to integrate pedagogical evaluators into multimodal learning analytics platforms \cite{daza2023edbb, daza2025smartevr}, combining instructional evaluation with biometric and physiological signals from students, such as indicators of cognitive load \cite{daza2024mebal2, daza2024deepface} and behavior cues \cite{Becerra2024, daza2024improveimpactmobilephones, becerra2025multimodal}. This multimodal integration aims to support more accurate modeling of student understanding by jointly analyzing instructional quality and learner state.

\begin{acknowledgments}
Support by projects: Cátedra ENIA UAM-VERIDAS en IA Responsable (NextGenerationEU PRTR TSI-100927-2023-2), M2RAI (PID2024-160053OB-I00, MICIU/FEDER), TRUST-ID (PID2025-173396OB-I00 MICIU/AEI and the EU) and PowerAI+ (SI4/PJI/2024-00062 Comunidad de Madrid and UAM). Javier Irigoyen is supported by a FPI fellowshop from MINECO/FEDER. 
\end{acknowledgments}

\newpage

\bibliography{sample-ceur}

\end{document}